\documentclass{article}
\usepackage[square,sort,comma,numbers]{natbib}

\usepackage[final]{neurips_2024}

\usepackage[utf8]{inputenc} 
\usepackage[T1]{fontenc}    
\usepackage{hyperref}       
\usepackage{url}            
\usepackage{booktabs}       
\usepackage{amsfonts}       
\usepackage{nicefrac}       
\usepackage{microtype}      
\usepackage{xcolor}         
\usepackage{comment}

\usepackage{tikz}
\usepackage{amsmath}

\usepackage{xcolor,color,xspace,enumerate,centernot,multirow,float,graphicx,
xcolor,caption,subcaption,textcomp,pgfplots,pgf-pie,tikz,listings,
comment,adjustbox,mdframed,changepage,algorithm,algorithmic}
\usepackage{hyperref}
\usepackage{filecontents}
\usepackage{xurl}
\usepackage{cryptocode} 
\usetikzlibrary{positioning, shapes.geometric, arrows}

\usepackage{amsmath}
\usepackage{amssymb}
\usepackage{xcolor}
\usepackage{msc}
\usepackage{graphicx}
\usepackage{booktabs}
\usepackage{array}
\usepackage{caption}
\usepackage{cleveref}

\title{The State of Julia for Scientific Machine Learning}

\author{%
  Edward Berman \thanks{Equal Contribution. Order determined by rock-paper-scissors, best 2 of 3.} \\ 
  Northeastern University\\
  Boston, MA 02115 \\
  \texttt{berman.ed@northeastern.edu} \\
   \And
  Jacob Ginesin \footnotemark[1]\\
  Northeastern University\\
  Boston, MA 02115 \\
  \texttt{ginesin.j@northeastern.edu} \\
}

\begin{document}

\maketitle

\begin{abstract}
  Julia has been heralded as a potential successor to Python for scientific machine learning and numerical computing, boasting ergonomic and performance improvements. Since Julia's inception in 2012 and declaration of language goals in 2017, its ecosystem and language-level features have grown tremendously. In this paper, we take a modern look at Julia's features and ecosystem, assess the current state of the language, and discuss its viability and pitfalls as a replacement for Python as the de-facto scientific machine learning language. We call for the community to address Julia's language-level issues that are preventing further adoption. 
\end{abstract}

\section{Introduction --- A Tale of Two Languages}

Historically, Python has been the dominant language for machine learning research, including in the sciences. Python is intuitive and has a rich ecosystem across all of the natural sciences. Yet, the rise of Python as the go-to language for machine learning has been in many ways unnatural. Python is a scripting language, extremely slow, and challenging to maintain. The advent of Julia was, in many ways, designed to address the limitations of Python. Julia boasts a suite of language-level features that are designed for intuitive and fast numerical computations.

Since Julia's introduction in 2012, the language has experienced steady growth among practitioners \cite{bezanson2017julia, julia_case_studies, julia_growth_stats_2024}. 
Given Julia's specific intent to solve problems in scientific machine learning, it's natural to ask why Julia hasn't challenged Python in popularity within the natural sciences. Some of this can be attributed to the \emph{momentum} Python has accumulated over the years within the community \cite{meyerovich2013empirical}. However, with the rapid growth of scientific machine learning in recent years, it is to be seen whether Julia will see broader adoption.

In this work, we explore the readiness of Julia as the de-facto tool for scientific machine learning. We focus not only on the wealth of libraries at Julia's disposal but also its performance, design philosophy, and overall ergonomics. A key theme in this work is that different programming languages provide different abstractions that significantly change the way a user interacts with a scientific machine learning problem (SMLP), and Julia provides a very different set of abstractions than what is seen in other ecosystems. In contrast to previous perspectives \cite{innes2018machine, gao2020julia}, we argue that while the Julia ecosystem provides a number of useful abstractions for SMLPs, the limitations are severe enough to prevent it from further adoption. We conclude by calling for the community to discuss and address Julia's language-level limitations.
 
\section{The Scientific Computing Ecosystem} \label{sec:ecosystem}
In this Section, we compare Julia's ecosystems with its contemporaries. We compare primarily with Python as it is the go-to choice for most researchers tasked with a SMLP. However, we recognize that languages such as C, C++, Fortran, and Cuda see similar usage at smaller scales. We point out examples of SMLPs that Julia is especially useful for throughout.

\paragraph{Linear Algebra} Arguably the most important part of the Julia ecosystem is the standard library for linear algebra. Fast numerical computations motivated Julia's inception. To achieve this, Julia employs a Just in Time (JIT) compiler to speed up its computation. The utility of the JIT compiler is seen across the language, but especially within its linear algebra library. For example, Julia is significantly faster than Python and about as fast as C in terms of random matrix multiplication and random matrix statistics \cite{bezanson2017julia}. Python matrix operations can be JIT compiled with Jax \cite{jax2018github}, however, the exact performance gain compared to C or Julia has yet to be properly assessed. Julia also has a plethora of easy to use functions for matrix factorization and libraries for working with sparse graphs that can make matrix operations even faster \cite{julia_sparsearrays}. This makes Julia the most compelling choice for SMLPs like SLAM \cite{rosen2019se, rosen2020certifiably}, where there is natural graph sparsity to exploit. There are also some widely adopted libraries for sparsity in C and C++ \cite{10.1145/3519024, 10.1145/3337792, 10.1145/2513109.2513116, Kolodziej2019, 10.1145/2049662.2049663, 10.1145/1114268.1114277, 10.1145/1824801.1824814, 10.1145/2491491.2491498, doi:10.1137/S0895479894246905, 10.1145/305658.287640, 10.1145/992200.992205, 10.1145/992200.992206, 10.1145/1024074.1024079, 10.1145/1024074.1024080, doi:10.1137/S0895479894278952, 10.1145/1024074.1024081, RENNICH2016140, doi:10.1137/S0895479897321076, doi:10.1137/S0895479899357346, doi:10.1137/S089547980343641X, 10.1145/1462173.1462176, 10.1145/1391989.1391995, 10.1145/3065870, 10.1145/2049662.2049670, FA02, 10.1145/3322125, xu2022pargemslr, Saad, falgout2020hypre}.

\paragraph{Constrained Optimization} In a similar flavor to numerical linear algebra, constrained optimization algorithms mesh well with Julia's strengths. Julia's constrained optimization algorithms are fast and mathematically sound, and in comparison to Python, much more abundant. 

Julia has the most sophisticated support for specifying optimization problems on manifolds.
Naive approaches to SMLPs do not account for inductive biases and physical constraints (e.g. \cite{berman2024efficientpsfmodelingshoptjl, Berman2024, rosen2020certifiably, rosen2019se, absil2008optimization, boumal2023intromanifolds}). One approach to endowing physical constraints into a problem is by specifying a manifold that the data lives on. This is implemented in Julia with Manopt \cite{Bergmann2022}. While there are Python and Matlab ports of Manopt \cite{boumal2014manopt,townsend2016pymanopt}, they offer only a strict subset of what the Julia library offers \cite{Bergmann2022}. The Python port in particular is very limited. There exists similar packages in R and C++ \cite{huang2018roptlib, martin2020manifoldoptim} but they are not actively maintained.

Manopt is designed to constrain iterative updates to stay on a specified manifold. However, often times constraints are specified softly via loss functions such as those in Physics Informed Neural Networks (PINNs). PINNs are able to take boundary and initial conditions into account when solving for a physical law specified as a differential equation. Julia handles this wonderfully, with Flux.jl \cite{innes2018flux, innes2018fashionable} and Lux.jl \cite{pal_2023} both providing neural network backends that are compatible with the DiffEqFlux.jl \cite{rackauckas2020universal} and NeuralPDE.jl \cite{NeuralPDE} packages. Even beyond neural methods, Julia has an extremely rich ecosystem for solving differential equations \cite{rackauckas2017differentialequations}, however this is out of the scope of this paper. Lux.jl in particular supports arbitrary types, making it composible with arbitrary ODE/PDE solvers and even other languages (more on this in $\S$ \ref{sec:dp}). In contrast, there are no de-facto solutions for physics informed machine learning in Python. That is, no existing solution accommodates neural networks specified in either TensorFlow \cite{TF}, Haiku \cite{haiku2020github}, or PyTorch \cite{paszke2019pytorch} simultaneously. As a result, we see standard architectures spread across different sub-ecosystems without any libraries available to bridge them together. For example, we see the canonical implementations for lagrangian neural networks \cite{cranmer2020lagrangian} and normalizing flows \cite{kidger2021on, halverson2024generalitypersistencecosmologicalstasis} implemented in Jax (possibly with Diffrax \cite{kidger2021on}), and the canonical implemenations of Kolmogorov Arnold Networks \cite{liu2024kan} and Deep Operator Networks \cite{li2024tutorials, lu2021learning} implemented in PyTorch. 

Julia rounds out its suite of constrained optimization libraries with GeometricFlux.jl \cite{tu2020geometricflux}. GeometricFlux.jl again builds on the Flux.jl framework with specific layers and functionality for the $5(+1)G$ fields, which includes \textbf{G}raphs and sets, \textbf{G}rids and Euclidean Spaces, \textbf{G}roups and Homogenous Spaces, \textbf{G}eodesics and Manifolds, \textbf{G}auges and Bundles, and \textbf{G}eometric Algebra \cite{bronstein2021geometric}. As with PINNs, these same functionalities appear scattered throughout different Jax and Torch libraries that are often incompatible with one another. This support makes Julia an attractive choice for working on SMLPs such as excited state molecular dynamics \cite{detanet, solvent2024}. There is also the GraphNeuralNetworks.jl library \cite{Lucibello2021GNN}, which is more actively maintained than GeometricFlux.jl, and has similar features.

\paragraph{Automatic Differentiation} Julia maintains several different packages for both forward mode and reverse mode automatic differentiation (AD) with ForwardDiff.jl \cite{revels2016forward} and Zygote.jl \cite{innes2018fashionable} respectively (see also \cite{enzyme,ChainRules,schafer2021abstractdifferentiation, reactantjl}). In other ecosystems, forward mode and reverse mode AD are typically handled within the same library, as is the case with PyTorch \cite{paszke2019pytorch} and Jax \cite{jax2018github}. In general, Julia's AD libraries are much more ambitious in scope, allowing for more overhead from the user at the potential cost of ease of use. We identify this as one of the few areas where there is a lot of friction when writing Julia code. One way in which this is realized is with custom gradient interventions and definitions. In Julia, macros for overriding and specifying differentiation rules are housed in many separate libraries, including Zygote.jl itself and the general purpose DiffRules.jl library \cite{Diffrules.jl}. Navigating the maze of different tools is often counter intuitive and involves much more domain knowledge than with other tools. In contrast, custom differentiation rules are very easy in Jax with the \textit{@custom\_JVP} and \textit{@custom\_VJP} macros.

Zygote.jl is most prominently used as the backend of the Flux.jl machine learning library. Additionally, the aforementioned automatic differentiation tools also enable Optim.jl \cite{Mogensen2018} for use with SymbolicRegressions.jl. In the space of symbolic regression, Julia has a clear advantage in terms of support. The most widely adopted library for symbolic regression is SymbolicRegressions.jl \cite{cranmer2023interpretable}. The Python alternative PySR is simply a wrapper for this library, which is in turn a dependency for PyTorch \cite{vid}. 

\paragraph{Probabilistic Programming} Julia boasts many Probabilistic Programming Languages (PPLs) built on top of it. These PPLs make SMLPs involving Bayesian statistics extremely simple to express, and include packages like SOSS.jl \cite{scherrer_soss_2024} and Turing.jl \cite{turing}. SOSS.jl uniquely provides a set of tools for working with measure-theoretic objects not seen in other languages \cite{measure_theory_jl_2024, Scherrer2022, scherrer_soss_2024}. Save for the measure theoretic support, a similar library also exists in Python with PyAutoFit \cite{Nightingale2021}. Additionally,  Jax \cite{jax2018github} with NumPyro \cite{phan2019composable, Foreman_Mackey_Intro_Numpyro} makes it exceptionally easy to specify probabilistic problems. 

\section{Design Philosophy and Ergonomic Machine Learning} \label{sec:dp}

Julia was founded on strong design ideals; in this section we discuss Julia's features and how they enable fluid development and solving of SMLPs.

\paragraph{Multiple Dispatch} As evidenced by \cite{bezanson2017julia}, Julia makes extensive use of multiple dispatch \cite{muschevici2008multiple}, a method to automatically choose function behavior based on input types.
In practice, this can make writing code much more ergonomic for the end user. As an example, the Python/C++ library for computing correlation functions, TreeCorr \cite{jarvis2004skewness}, has $18$ different subclasses that the user must remember in order to correlate different kinds of objects. All of these subclasses are prefixed with single letters, such as \textsc{KVCorrelation} for scalar-vector correlations and \textsc{VVCorrelation} for vector-vector correlations. Needless to say, it can be difficult to keep track of all $18$ combinations, but with Julia's multiple dispatch, we can safely define the same function $18$ times for the different types of inputs without having to worry about specifying which one to call in practice. This is the approach also taken by CosmoCorr.jl \cite{CosmoCorr}. Other libraries that make extensive use of multiple dispatch include JuMP.jl \cite{dunning2017jump}, ForwardDiff.jl \cite{revels2016forward}, and Julia's standard library.

\paragraph{Composition vs. Inheritance} Julia's design philosophy and style guides revolve around the idea of highly composable interfaces. This naturally plays off of the use of multiple dispatch: functions are not attached to a single struct and instead are declared globally, potentially multiple times for different types. An example of this can be seen with Flux.jl \cite{innes2018flux, innes2018fashionable}. A neural network in Flux.jl is just a chain of functions that can act on different types. This makes Flux.jl much more ergonomic as it abstracts away the different input types you may provide to a neural network, eliminating the need to extend a module and format the input data in a certain way (i.e. as one would do with PyTorch \cite{paszke2019pytorch}). It is for this reason that Flux.jl boasts that it is the library that \emph{doesn't make you tensor}. The emphasis on composition is also useful for package management. Since there is very little shared state between different user defined structs due to multiple dispatch, packages are less likely to conflict. Even when there are conflicts, Julia's package manager Pkg.jl \cite{Pkg.jl, phanson_2024} is much easier to use. Python's package management on the contrary, which depends on Python virtual environment management with Pip or Conda, is notoriously error prone.

\paragraph{Two Language Problem} The two language problem states that a programming language is either fast or easy to use, but not both. Thus, we often write performance critical code in C or C++ and call it from Python. This makes projects difficult to maintain as there are now multiple environments the user needs to keep track off. This also makes reproducibility and contribution more difficult as it requires everyone in the scientific community to know a second,  low level technical language\footnote{or, at least how to compile it with make or cmake, given the project doesn't provide a dockerfile or similar.}. The most glaring example of this is PyTorch \cite{paszke2019pytorch}, which is a Python interface to a C++ backend. However, even libraries that provide much simpler functionality than a neural network have this issue. For example, TreeCorr also has a Python API for functions written in C++ \cite{jarvis2004skewness}. Julia claims to solve the two language problem, allowing both flexible prototyping and deep performance optimization within the same language \cite{bezanson2017julia}. 
For example, the Flux.jl \cite{innes2018flux, innes2018fashionable} and CosmoCorr.jl \cite{CosmoCorr} libraries are able to efficiently train neural networks and estimate correlation functions all with high-level Julia code. In general, the interplay between prototyping and optimizing is a core design pattern when implementing solutions for SMLPs, and Julia's design uniquely facilitates this interplay. 
 
\section{Limitations of Julia for Scientific Machine Learning} \label{sec:limitations}

In this section, we outline the limitations we find most severe in Julia for solving SMLPs.

\paragraph{Lack of Software Engineering Features} In stark comparison with Python, Julia lacks many software engineering-oriented language and ecosystem features. Primarily, Julia has significantly less mature testing infrastructure than Python; the Unittest and Pytest Python libraries are much more robust than Julia's built in testing library. Additionally, Julia has virtually no support for more advanced testing methods such as property-based testing, symbolic execution, and contract-based testing, of which are universally employed by Python's large-scale numerical methods and machine learning libraries such as Pandas \cite{mckinney2011pandas}, NumPy \cite{harris2020array}, SciPy \cite{virtanen2020scipy}
, SymPy \cite{meurer2017sympy}, Scikit-Learn \cite{pedregosa2011scikit}, Jax \cite{jax2018github}, Tinygrad \cite{tinygrad}, and Cupy \cite{nishino2017cupy} through the Hypothesis \cite{hypothesisworks_hypothesis_2024}, Crosshair \cite{pschanely_crosshair_2024}, and Deal \cite{life4_deal_2024} libraries. Julia also lacks an actively maintained static type checker \cite{pyright_docs, shivarpatna2024typeevalpy, mypy}. More rigorous testing methods are especially important in scientific computation, where precision and correctness of implementations of algorithms is paramount. In being able to deeply test properties of an algorithm, we can better reconcile epistemic (model) uncertainties \cite{osband2023epistemic} from aleatoric (data) uncertainties. 

\paragraph{Complex Debugging Messages} On a language level, Julia's error messages have continually been an unaddressed pain point in the community. The complexity of multiple dispatch, as well as the notoriously long stack traces, make digesting Julia's error messages a tedious process. Additionally, a recent Julia Hub survey showed that Julia programmers cited complex debugging messages as one of the biggest technical hurdles for the language \cite{juliaconsurvey, juliahubsurvey}.

\paragraph{Limited Industry Adoption} While Julia may be gaining popularity in academic circles, machine learning research has a heavy industry presence unseen in many other disciplines. As such, open source contributions from industry is a huge factor in the overall adoption of a language or library. A key limitation of Julia is its lack of support from industry giants. Julia's biggest competitor is arguably Jax \cite{jax2018github}, which is maintained by Google Deepmind --- we see Jax as something that treats the symptoms of Python's problems, the most significant of which is its performance. Moreover, companies like HuggingFace have designed their API's for sharing model weights and datasets exclusively around the Python machine learning libraries, making it much more difficult for Julia users to share their scientific models with one another and necessitating the use of Python wrappers (e.g. \cite{cheng_2024_transformersjl}). While Julia has a growing number of industry backers \cite{julia_case_studies}, there is a clear resource advantage for the Python ecosystems. Supporting this, the 2024 Julia Hub survey \cite{juliaconsurvey,juliahubsurvey} found 64\% of users thought one of the biggest non-technical problems with the language was ``there are not enough Julia users in my field or industry,'' a jump of $\approx 6\%$ from 2023. The 2024 survey also found that while $71\%$ of respondents use Julia for research, only $16\%$ reported ``I use Julia in production for a business critical task'' \cite{juliaconsurvey,juliahubsurvey}. Industrial applications require robust testing and interoperability with existing tools, supporting the idea that the limitations laid out in this section are not at all unrelated.

\paragraph{Poor Interoperability} Despite the annoyances of the two language problem laid out in $\S$ \ref{sec:dp}, users will inevitably need to use multiple languages until scientists start writing their domain specific applications in Julia. While calling Python, R, C, and C++ functions in Julia is relatively easy with the PythonCall.jl \cite{PythonCall}, PyCall.jl \cite{pycalljl}, RCall.jl \cite{lai_2024}, and the Julia standard library function \emph{ccall} \cite{julia_calling_c_fortran}, the converse is often extremely difficult and unintuitive as the user will need to explicitly manage Julia's context and the passing of complex datatypes between languages. While \cite{PythonCall} handles the Julia to Python conversion most graciously, there is still much to be desired in calling Julia from other languages.

\section{Conclusion and Call to Action} \label{sec:conclusions}

Not only does Julia match and in many ways exceed Python with its suite of general purpose scientific computing tools, but it also provides many more specific tools for solving SMLPs unseen in any other language. This is particularly true for constrained optimization problems that naturally arise in the natural sciences. Unfortunately, this has proven to be not enough for Julia to gain widespread adoption. While Julia has been slowly gaining traction, in comparison to the growth of scientific machine learning as a whole, Julia's growth has been quite slow.  

Julia's lack of software engineering-centric features, lackluster debugging infrastructure, subpar industry adoption, its ``rivalry'' with Jax \cite{kidger2024jaxvsjulia}, and insufficient interopt all bottleneck further adoption. However, none of these limitations in isolation seem like dealbreakers. On paper, Julia \emph{should} compete with Python. So again we ask, why doesn't it?

In the seminal paper, ``Julia: A Fresh Approach to Numerical Computing'' \cite{bezanson2017julia}, the initial vision for Julia is summarized in three points. $1)$ A programming language can be high-level, dynamic, and fast. $2)$ A programming language can be used for both scripting and deployment. $3)$ A programming language should supply a mechanism to easily abstract away unnecessary detail typically left to ``the experts.'' Arguably, Julia today has addressed all three of these points and matured with respect to them. Yet, Julia remains relatively niche despite the growing potential user base.

In present year, Julia users have mostly moved away from improving the Julia language and are now more  focused on developing libraries for specific projects. With love, we argue the Julia community has moved on too quickly. The state of machine learning is rapidly changing, and 
Julia has the potential to address many of the pain points in the community. However, if we do not address problems at the language level, we are in danger of repeating many of the mistakes of Python.

We believe the current state of Julia lacks vision. Julia needs a new constitution: a set of concrete goals for improvement, adoption, and outreach. While Python has a clear list of future goals \cite{peps_python}, and individual ecosystems \emph{within} Julia such as SciML have roadmaps \cite{sciml_roadmap}, the Julia language itself has nothing but surface-level GitHub issues. It is only once we map out and solve these language-level issues that we can then ask if Julia is capable of succeeding Python as the de-facto language for scientific machine learning.

We ask the Julia and scientific machine learning communities: what is the future for Julia?

\ack

The authors acknowledge helpful discussions with the Julia community members on the Julia Discourse and Slack. The authors would also like to thank Beck LaBash, Jacob S. Zelko, and Arjun Guha for helpful discussion and comments. 

\bibliographystyle{plain}
\bibliography{main}

\appendix

\end{document}